\documentclass[journal]{IEEEtran}

\usepackage{amssymb}
\usepackage{latexsym}

\usepackage{url}
\usepackage{xcolor}
\usepackage{color,soul}
\usepackage{caption,subcaption,booktabs}

\usepackage{amsmath}
\usepackage{mathtools}
\usepackage{enumitem}

\usepackage{dblfloatfix}
\usepackage{tikz}       
\usepackage{nicematrix} 
\usepackage{framed,multirow}

\usepackage{lipsum}
\usepackage[switch]{lineno}
\usepackage{tcolorbox}
\usepackage[colorlinks=true,linkcolor=red]{hyperref}

\usepackage[labelsep=period,labelfont=bf]{caption}
\usepackage[nameinlink]{cleveref}  
\crefname{figure}{Fig.}{\textbf{Figure.}}
\crefname{equation}{Eq.}{\textbf{Eq.}}
\crefname{table}{Table}{\textbf{Table.}}
\crefname{section}{Section}{\textbf{Section}}

\definecolor{newcolor}{rgb}{.8,.349,.1}

\hypersetup{
	colorlinks,
	linkcolor=[rgb]{1.0, 0.0, 0.0},
	citecolor=[rgb]{1.0, 0.2, 0.2},
	urlcolor =[rgb]{0.0, 0.0, 0.8}
}

\begin{document}

\title{IMPaSh: A Novel Domain-shift Resistant Representation for Colorectal Cancer Tissue Classification}

\author{~Trinh~Thi~Le~Vuong$^{1,*}$, ~Quoc~Dang~Vu$^{2,*}$\\ ~Mostafa~Jahanifar$^{2}$, ~Simon~Graham$^{2}$ \\
~Jin~Tae~Kwak$^{1,+}$, ~Nasir~Rajpoot$^{2,+}$ \\
\{trinhvg, jkwak\}@korea.ac.kr \\
\{quoc-dang.vu, mostafa.jahanifar, simon.graham, n.m.rajpoot\}@warwick.ac.uk \\ 
$^*$ Joint First Authors Contributed Equally. \\
$^+$ Joint Last Authors Contributed Equally.
\thanks{T.T.L.Vuong and J.T.Kwak are from School of Electrical Engineering, Korea University, Seoul, Korea}
\thanks{Q.D.Vu, M.jahanifar, S.Graham and N.Rajpoot are from the Tissue Image Analytics Centre, Department of Computer Science, University of Warwick, UK}
}

\maketitle

\begin{abstract}
The appearance of histopathology images depends on tissue type, staining and digitization procedure. These vary from source to source and are the potential causes for domain-shift problems. Owing to this problem, despite the great success of deep learning models in computational pathology, a model trained on a specific domain may still perform sub-optimally when we apply them to another domain. To overcome this, we propose a new augmentation called PatchShuffling and a novel self-supervised contrastive learning framework named IMPaSh for pre-training deep learning models. Using these, we obtained a ResNet50 encoder that can extract image representation resistant to domain-shift. We compared our derived representation against those acquired based on other domain-generalization techniques by using them for the cross-domain classification of colorectal tissue images. We show that the proposed method outperforms other traditional histology domain-adaptation and state-of-the-art self-supervised learning methods.  Code is available at: \url{https://github.com/trinhvg/IMPash}.

\end{abstract}


\IEEEpeerreviewmaketitle

\section{Introduction}
Although Deep learning (DL) models have been shown to be very powerful in solving various computational pathology (CPath) problems \cite{srinidhi2021deep}, they can be very fragile against the variations in histology images \cite{footereet}. One of the main challenges in CPath is \textit{domain-shift} where the distribution of the data used for training and testing of the models varies significantly. There are many sources that can cause domain-shift in CPath, such as variation in sample preparation and staining protocol, the colour palette of different scanners and the tissue type itself.

In CPath, there have been several efforts attempting to solve the domain-shift problem by using stain normalization, stain-augmentation, domain adaptation or domain generalization techniques \cite{jahanifar2021midog,koohbanani2021self}. The objective of stain normalization (SN) is to match the colour distribution between the training and testing domains  \cite{vahadane2016structure,alsubaie2015discriminative}. On the other hand, stain augmentation (SA) tries to artificially expand the training set with stain samples that could potentially be within the unseen testing set. This is often achieved by randomly changing the stain/colour information of source images \cite{pocock2021tiatoolbox}. In certain problems, radically altering the image colour from its usual distribution may also be beneficial, such as by using medically-irrelevant style transfer augmentation\cite{yamashita2021learning}.

On the other hand, both domain adaptation and domain generalization families try to directly reinforce the model's ability to represent images, such as via the loss functions, to achieve robustness across training and testing domains. The former relies on data from the unseen domain, whereas the latter does not. And due to such reliance on the data of target domains, domain adaptation techniques are thus also highly dependent on the availability and quality of curated data in such domains. Abbet et \textit{al.} \cite{abbet2021self} recently proposed a training scheme that we can consider a prime example of a domain adaptation technique. In particular, they employed in-domain and cross-domain losses for colorectal tissue classification.

As for domain generalization, recent proposals mostly focus on pre-training models. In fact, self-supervised algorithms such as MoCoV2 \cite{chen2020improved} or Self-Path \cite{koohbanani2021self} are particularly attractive because they can leverage a huge number of unlabelled histology images. However, in order to effectively train such a model based on self-supervised contrastive learning (SSCL) methods, a careful selection of augmentations and their corresponding parameters is of crucial importance.

In this research, we propose training a DL-based model in self-supervised contrastive learning (SSCL) manner so that the resulting model can extract robust features for colorectal cancer classification across \textit{unseen} datasets. We propose a new domain generalization method inspired by \cite{tian2020makes,chen2020improved} that does not rely on data in the domains to be evaluated. Our contributions include:

\begin{itemize}
\item We propose PatchShuffling augmentation to pre-train the model such that they can extract invariant representation.
\item We propose a new SSCL framework that combines InfoMin\cite{tian2020makes} augmentations and PatchShuffling for model pre-training called \textbf{IMPaSh} based on contrastive learning with momentum contrast.
\item We provide a comparative evaluation to demonstrate the effectiveness of IMPaSh.
\end{itemize}

\section{The Proposed Method}

\begin{figure*}[!t]
\centering
\includegraphics[width=1.\textwidth]{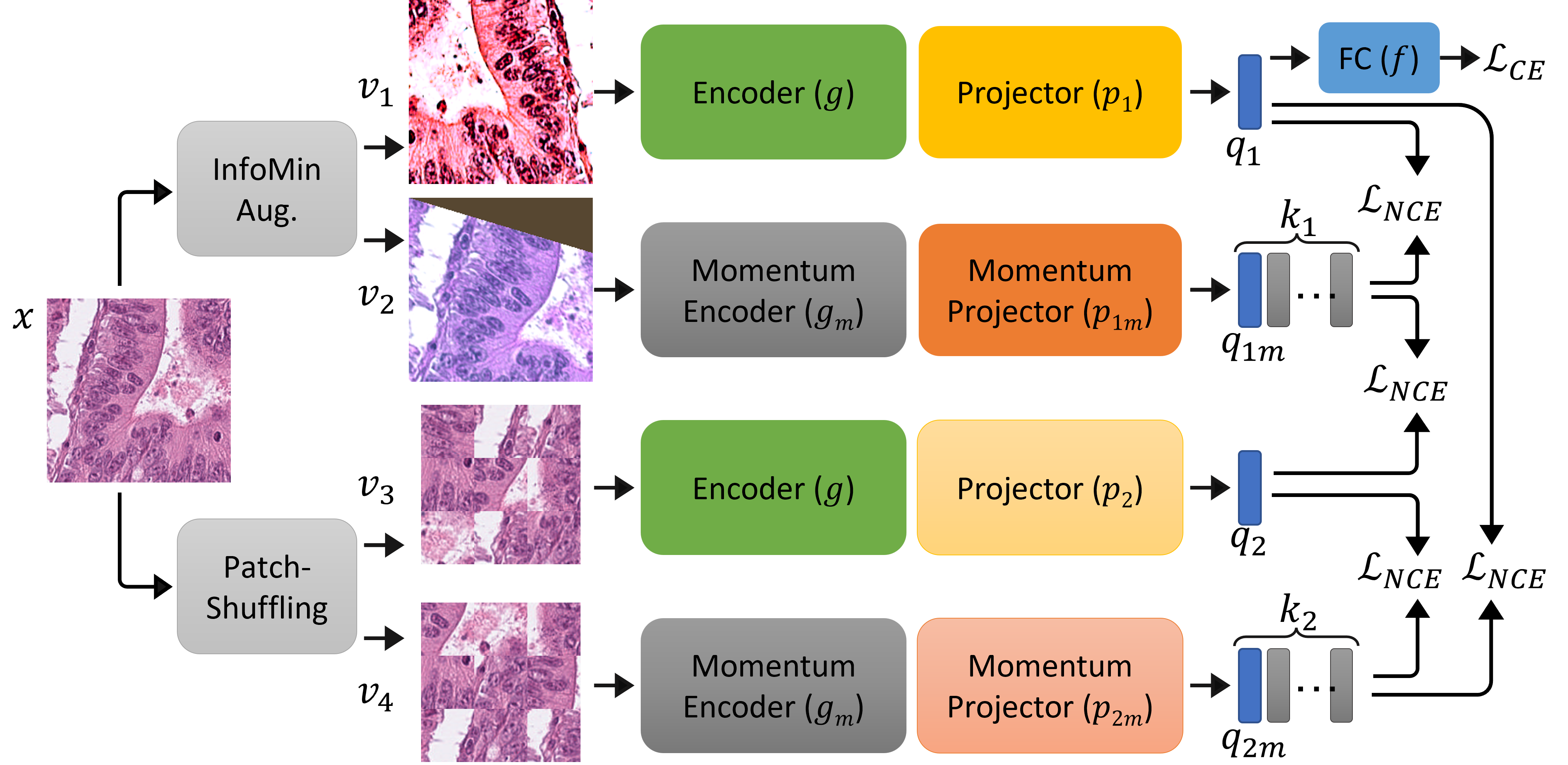}
\caption{Overview of the proposed self-supervised contrastive learning method. \textit{FC} denotes a fully connected layer that acts as a classifier (i.e., the classification head). In the figure, encoders that are of the same color have their weights shared.}
\label{fig1}
\end{figure*}

\subsection{Self-supervised contrastive learning}

An overview of our proposed IMPaSh method is presented in Fig. \ref{fig1}. As can be seen in the figure, encoders and projectors that are of the same color have their weights shared. The proposed method includes two types of augmentation: InfoMin \cite{tian2020makes} and our proposed PatchShuffling (Section \ref{PatchShuffling}). For each approach, an input image $x$ is augmented twice, resulting into 4 different views $\{v_1, v_2, v_3, v_4\}$ of $x$. An encoder $g$ and two multilayer perceptron (MLP) projectors, called $p_1$ and $p_2$, are applied on $v_1$ and $v_3$ to respectively extract feature vector $q_1$ and $q_2$ both in $\mathbb{R}^{128}$. As for $v_2$ and $v_4$, in a slightly different manner, their corresponding feature vector $q_{1m}$ and $q_{2m}$ are extracted by the \textit{momentum} version of $g$ and $p$, which we subsequently denote as $g_m$, $p_{1m}$ and $p_{2m}$. These $q_{1m}$ and $q_{2m}$ features are further used to update 2 queues of feature vectors $k_1$ and $k_2$.


For the SSCL task, query $q_1$ and query $q_2$ are then respectively compared to $k_1$ and $k_2$ to allow the model to learn the similarity between the different views. Training is thereby achieved by optimizing the InfoNCE loss ($\mathcal{L}_{NCE}$) \cite{oord2018representation}. Specifically, new $q_1$ and $q_2$ treat existing features that are from different images in the queue as negative samples for optimization.

For the transfer learning task, we first freeze the encoder $g$ and projector $p_1$ and then add another classifier on top of projector $p_1$. We further describe the implementation of the PatchShuffling augmentation and momentum contrast in the following sections.

\subsection{Learning Pretext-Invariant Representation }
\label{PatchShuffling}

Self-supervised learning endows the model with the ability to extract robust representation by maximizing the similarity across variations of the same image. Traditionally, such variations are obtained by covariant transformations like translation or rotation \cite{gidaris2018unsupervised,goyal2019scaling}. However, recent investigations have shown that invariant transformations, such as Jigsaw Puzzle Solving \cite{noroozi2016unsupervised} and PIRL \cite{misra2020self}, are more powerful. In the case of PIRL in particular, the model is forced to learn a representation of the image based on its constituent smaller patches, regardless of their positions. In this paper, we propose PatchShuffling based on PIRL for learning invariant representation and pre-train the model using not only PatchShuffling but also InfoMin.

\subsubsection{PatchShuffling:} We first randomly crop a portion of from the image $x$ such that its size is around $[0.6, 1.0]$ of the original image area. We then resize it to 255 $\times$ 255 pixels and randomly flip the image using the settings in \cite{misra2020self,tian2020makes}. Afterward, we divide the image into a grid of 3 $\times$ 3 cells each which occupies 85 $\times$ 85 pixels. We further crop each cell randomly to 64 $\times$ 64 pixels and then randomly re-assemble them back to an image of 192 $\times$ 192 pixels. In comparison to PIRL, which first extracts the patch feature and then shuffles the the placement of the features within the original image, PatchShuffling only performs the shuffling on the original image itself.

\subsubsection{InfoMin:} We construct views $v_1$ and $v_2$ of a given image $x$ by using the augmentation setup in \cite{tian2020makes}. In particular, InfoMin augmentation is specifically designed so that the mutual information of the original image and the augmented images is as low as possible while keeping any task-relevant information intact.

\subsection{Momentum Contrast}


In contrastive learning, the most common approach for end-to-end learning is by using only the sample in the current iteration \cite{chen2020simclr}. However, in order to obtain a good image representation, contrastive learning requires a large set of negative samples. Thus a large batch size is required for the training processes (i.e., high GPU memory demand). To handle this memory problem, Wu et \textit{al.} \cite{wu2018unsupervised} proposed a memory bank mechanism that stores all the features obtained from previous iterations. Then, from this memory bank, a set of negative samples are randomly selected for the current training iteration. As a result, a large number of samples can be obtained without relying on back-propagation, which in turn dramatically decreases the required training time. However, because the selected samples may come from different training iterations (i.e., from vastly different encoders), there may exist a large discrepancy between them which can severely hinder the training process. To alleviate this problem, MoCo \cite{he2020momentum} introduced momentum contrast which allows the construction of a consistent dictionary of negative samples in near linear scaling. Inspired by this, to best exploit contrastive learning, we utilize momentum contrast for both InfoMin and PatchShuffling by constructing two dedicated momentum branches as introduced in Fig. \ref{fig1}.

The first momentum contrast branch encodes and stores a dictionary of image representations from an image augmented based on InfoMin. Meanwhile, the second one handles the representation of PatchShuffling. Parameters of these momentum encoders and projectors are updated following the momentum principle.
Formally, we denote the parameters of the momentum branch $\{g_{m}, p_{1m}, p_{2m}\}$ as $\theta_m$ and $\{g, p_{1}, p_{2} \}$ as $\theta_q$, we update $\theta_m$ by:
\begin{equation}
\theta_m \leftarrow \alpha \theta_m + (1-\alpha)\theta_q.   
\end{equation}
Here, $\alpha$ is a momentum coefficient to condition the training process to update $\theta_m$  more than $\theta_q$. We empirically set $\alpha = 0.9999$ to the value as used in MoCo \cite{he2020momentum}.

\subsubsection{Loss function:}
Our proposed loss function is an extended version of InfoNCE loss \cite{oord2018representation}. In essence, the loss maximizes the mutual information between positive pair obtained from an encoder and its momentum version. At the same time, the loss also tries to minimize the similarity in the representation of the current view of the image compared to other K = 65536 negative samples from the momentum encoder.

We denote the encoded query as $q$, where $q_1 = p(g(v_1))$, and $q_2 = p(g(v_3))$, and denote the set of keys from momentum branch as $k$, where $k_1 = p_m(g_m(v_2))$, and $k_2 = p_m(g_m(v_4))$. Objective function $\mathcal{L}_{NCE}$ for each pair of query $q$ and queue $k$ is defined as
\begin{equation}
\mathcal{L}_{NCE}(q, k) = -\mathbb{E} \left[ \log\frac{\exp(q_i \cdot k_i /\tau)}{\sum_{j=1}^K\exp(q_i \cdot k_j/\tau)} \right]
\end{equation}
where the temperature hyper-parameter $\tau=0.07$. In summary, the objective function for our contrastive learning framework is as follows,
\begin{equation}
\begin{split}
\mathcal{L}_{NCE} = & \mathcal{L}_{NCE}(q_1, k_1) + \mathcal{L}_{NCE}(q_1, k_2) \\ 
& + \mathcal{L}_{NCE}(q_2, k_1) + \mathcal{L}_{NCE}(q_2, k_2)
\end{split}
\end{equation}

\subsection{Transfer learning task}
After pre-training the self-supervised task, we obtain encoder $g$ and projector $p$. Instead of discarding the momentum branch and all projectors, we keep and freeze both the projector $p$ and the encoder $g$ to embed the input image to 128-dimensional features. Then, in a supervised manner, we can train another classifier $f$ on top of these features using cross-entropy loss ($\mathcal{L}_{CE}$) and the labels from the corresponding training dataset.

\section{Experiment}

\subsection{The Datasets}

\begin{figure*}[!t]
\centering
\includegraphics[width=1.\textwidth]{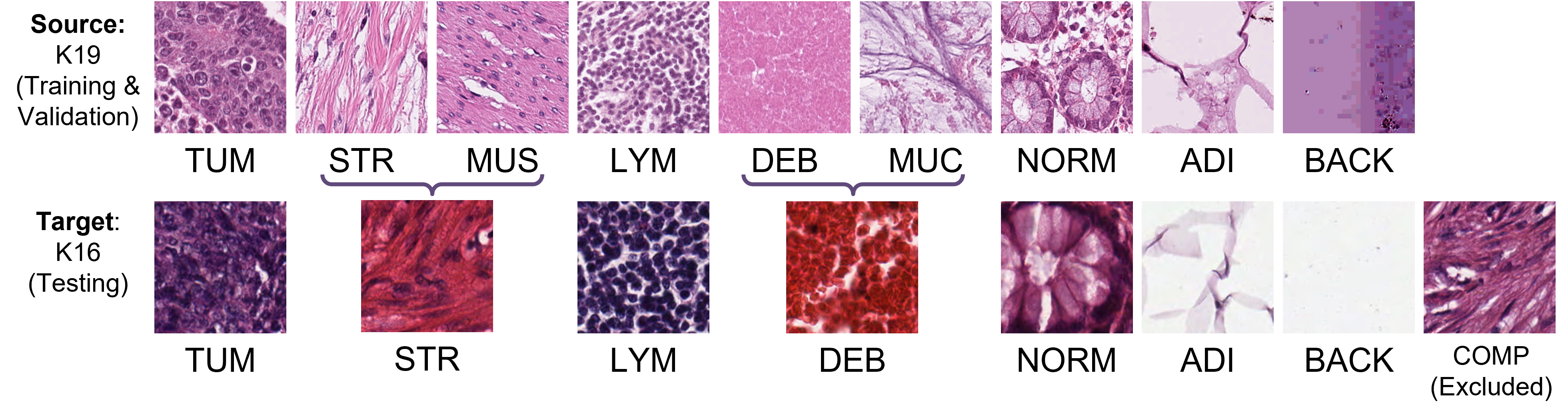}
\caption{Example images from the two used datasets of colorectal histology images: K19 and K16. Since these two datasets have different class definitions, stroma/muscle and debris/mucus in K19 are grouped into the stroma (STR) and debris (DEB) in K16. On the other hand, complex stroma (COMP) in K16 is excluded from the analyses.}
\label{fig_dataset}
\end{figure*}

We employed two publicly available datasets of colorectal histology images to evaluate our method: 1) K19 \cite{kather2019predicting} dataset, which includes 100,000 images of size 244$\times$224 pixels from 9 tissue classes as the source domain, and 2) K16 \cite{kather2016multi} containing 5,000 images of size 150$\times$150 pixels from 8 classes as the target domain. Example images from these two datasets are shown in Fig. \ref{fig_dataset}. Since these two datasets have different class labels, we followed \cite{abbet2021self} to group 9 classes from the training dataset (K19) into 7 classes that are best matched to the 7 classes in the test set (K16). In particular, stroma/muscle and debris/mucus are grouped as stroma (STR) and debris (DEB), respectively. Additionally, for K16, we excluded 625 ``complex stroma" images due to the lack of that group in the training domain, leaving us with a test set of 4,375 images. In total, the 7 classes that we evaluate are: adipole (ADI), background (BACK), debris (DEB), lymphocyte (LYM), normal (NORM), stroma (STR) and tumour (TUM).

\subsection{Experimental Settings}
In this study, we adopted ResNet-50 \cite{he2016deep} feature extractor as our backbone network (i.e the encoder). All the projectors in the self-supervised training stage consist of 2 fully-connected layers. Meanwhile, the classifier $f$ in the transfer-learning stage consists of only one fully-connected layer. Both pre-trained encoder $g$ and classifier $f$ were trained using 4 GPUs with a batch size of 256 and optimized with SGD default parameters. We trained the encoder $g$ following MoCo settings which utilized 65,536 negatives samples. The encoder was trained for 200 epochs with an initial learning rate of 0.03 and decayed based on a cosine annealing schedule. On the other hand, the linear classifier $f$ was trained for 40 epochs. Its learning rate started at 30 and then reduced by 5 at epoch 30. For evaluating the performance, we measured the accuracy (Acc), recall (Re), precision (Pre) and F1 of each class, then we took the averaged and reported the results.

\subsection{Comparative Experiments}
We compare our methods with several existing domain generalization methods.
\subsubsection{Domain-specific methods} 1) (SN Macenko) \cite{macenko2009method} is a stain normalization method proposed by Macenko, which is used to stain normalized the whole training dataset. 2) SN Vahadane \cite{vahadane2016structure} is another stain normalization method aiming to preserve histology images' structure.
\subsubsection{Self-supervised methods} 1) InsDis \cite{wu2018unsupervised} formulate the supervised problem as instance-level discrimination, which stores feature vectors in a discrete memory bank and directly compares distances (similarity) between instances. 2)PIRL \cite{misra2020self} is a self-supervised method that employs the pre-text task representation, and memory bank \cite{wu2018unsupervised} to store the negative sample in the self-supervised contrastive task. There are two views in PIRL; one is the original image while the other is cropped into nine patches, encoding these nine patches into nine 128 dimension vectors and then concatenating them in random order. 3) MocoV2  \cite{chen2020improved} is similar to InsDis in constructing multiple views, but in MoCoV2, negative samples are stored in the momentum updated manner. 4) InfoMin \cite{tian2020makes} is a combination of InfoMin augmentation and PIRL \cite{misra2020self}. InfoMin constructs two views while PIRL constructs one more view; there is only one momentum encoder that constructs a dictionary of negative samples, i.e., negative samples only come from the InfoMin augmentation.

\subsection{Ablation Study}

We conducted experiments to investigate the contribution of the projection heads and the momentum encoders on the model's overall performance when testing on a different domain. In addition to that, we also compared the benefits of shuffling only on the original images (PatchShuffling) against doing the shuffling in feature space (PIRL).

\section{Results and Discussion}

\begin{table*}[!t]
\centering
\caption{Results of the domain generalization experiments between a different source domain (K19) and target domains (K16) using various domain-adaptation techniques}
\label{tab1}
\begin{tabular}{c|c|c|c|c|c}
\toprule
\textbf{Method} & \textbf{Training Set} &	\textbf{Acc} &	\textbf{Re} &	\textbf{Pre} &	\textbf{F1} \\
\midrule
ImageNet - Upper Bound & K16 (Target) &	0.942 &	0.942 &	0.941 &	0.941 \\
ImageNet - Lower Bound &	K19 &	0.654 &	0.654 &	0.741 &	0.626 \\
SN Macenko \cite{macenko2009method} &	K19 &	0.660 &	0.660 &	0.683 &	0.645 \\
SN Vahadane \cite{vahadane2016structure} &	K19 &	0.683 &	0.683 &	0.696 &	0.656 \\
\midrule
InsDis \cite{wu2018unsupervised} &	K19 &	0.694 &	0.694 &	0.766 &	0.659 \\
PIRL \cite{misra2020self} &	K19 &	0.818 &	0.818 &	0.853 &	0.812 \\
MocoV2  \cite{chen2020improved} &	K19 &	0.675 &	0.675 &	0.816 &	0.642 \\
InfoMin \cite{tian2020makes} &	K19 &	0.750 &	0.750 &	0.824 &	0.752 \\
\midrule
IMPaSh (Ours) & K19 &	\textbf{0.868} &	\textbf{0.868} &	\textbf{0.887} &	\textbf{0.865} \\
\bottomrule
\end{tabular}
\end{table*}

\begin{table*}[!t]
\centering
\caption{Ablation results when we trained each component of IMPaSh on K19 (the source domain) and then independently tested them on the K16 (the target domain). \textit{ME} denotes using extra Momentum Encoder while \textit{Head} denotes using an additional projector}
\label{tab2}
\begin{tabular}{c|c|c|c|c|c|c}
\toprule
\textbf{Method} & \textbf{Head} & \textbf{Add ME} &	\textbf{Acc} &	\textbf{Re} &	\textbf{Pre} &	\textbf{F1} \\
\noalign{\smallskip}
\toprule


\begin{tabular}{@{}c@{}}InfoMin + PIRL\end{tabular} & \checkmark & & 0.862 &	0.862 &	0.875 &	0.859 \\
\begin{tabular}{@{}c@{}}InfoMin + PIRL\end{tabular} & \checkmark & \checkmark & 0.838 &	0.838 &	0.867 &	0.836 \\

\midrule

\begin{tabular}{@{}c@{}}InfoMin + PatchShuffling \\ (IMPaSh --- ours)\end{tabular} &	\checkmark & &	0.855 &	0.855 &	0.870 &	0.852 \\
\begin{tabular}{@{}c@{}}InfoMin + PatchShuffling \\ (IMPaSh --- ours)\end{tabular} &	\checkmark &	\checkmark  &	\textbf{0.868} &	\textbf{0.868} &	\textbf{0.887} &	\textbf{0.865} \\
\bottomrule
\end{tabular}
\end{table*}

Table \ref{tab1} reports the performance of ResNet50 when training and validating on K19 and testing on the unseen dataset K16. It is clear that using pre-trained ImageNet weights for the ResNet50 feature extractor and training the classification head $f$ on a dataset is enough to achieve good results on the test data from the same domain with a 0.942 accuracy (\textit{ImageNet - Upper Bound}). However, when we trained the same model on the K19 dataset and tested it on the K16 dataset, it performed poorly, with a drop of more than 25\% in accuracy (\textit{ImageNet - Lower Bound}). Despite the simplicity of the task at hand and the well-known capacity of deep neural networks, these results demonstrate that when a testing dataset is of different distribution compared to the training set, the deep learning model can still fail to generalize. As such, our main target is to identify techniques that can improve the performance of any model trained on K19 (\textit{ImageNet - Lower Bound}) to the extent that they can be comparable to those trained directly on K16 (\textit{ImageNet - Upper Bound}). 

Given the fact that K16 and K19 were obtained via different protocols, i.e., their color distribution is different, the above premise and results suggest that utilizing simple domain-specific adaptation techniques such as stain normalization could improve the model performance. In this work, we evaluated this hypothesis by employing the Macenko method  \cite{macenko2009method} and Vandadane method \cite{vahadane2016structure}. From Table \ref{tab1}, we demonstrate that such conjecture is plausible as the model performance on F1 scores on the unseen K16 dataset was improved by about 2-3\%.

We further evaluated the effectiveness of domain generalization techniques. Table \ref{tab1} shows that not all self-supervised learning approaches are noticeably superior to the more simple stain normalization methods. In fact, out of all comparative self-supervised learning methods, only PIRL and InfoMin achieved higher performance compared to the baseline ImageNet encoder and stain normalization techniques with more than 10\% difference in F1 and at least 7\% difference in accuracy.

From Table \ref{tab1}, it can also be observed that our proposed IMPaSh outperformed all other domain-specific techniques and state-of-the-art SSCL approaches such as stain normalization and MoCoV2 \cite{chen2020improved,tian2020makes,misra2020self,tian2020contrastive,wu2018unsupervised}. In particular, the F1 score on the unseen K16 dataset increases by 24\% using our proposed SSCL approach compared to using ImageNet weights without using any stain normalization during test time. In comparison with MoCoV2 \cite{chen2020improved} which has a single momentum branch and PIRL \cite{misra2020self} which only uses Jigsaw Puzzle Solving augmentation, our method respectively achieved 22\% and 5\% higher F1-score. 

\begin{figure*}[!t]
\centering
\includegraphics[width=\textwidth]{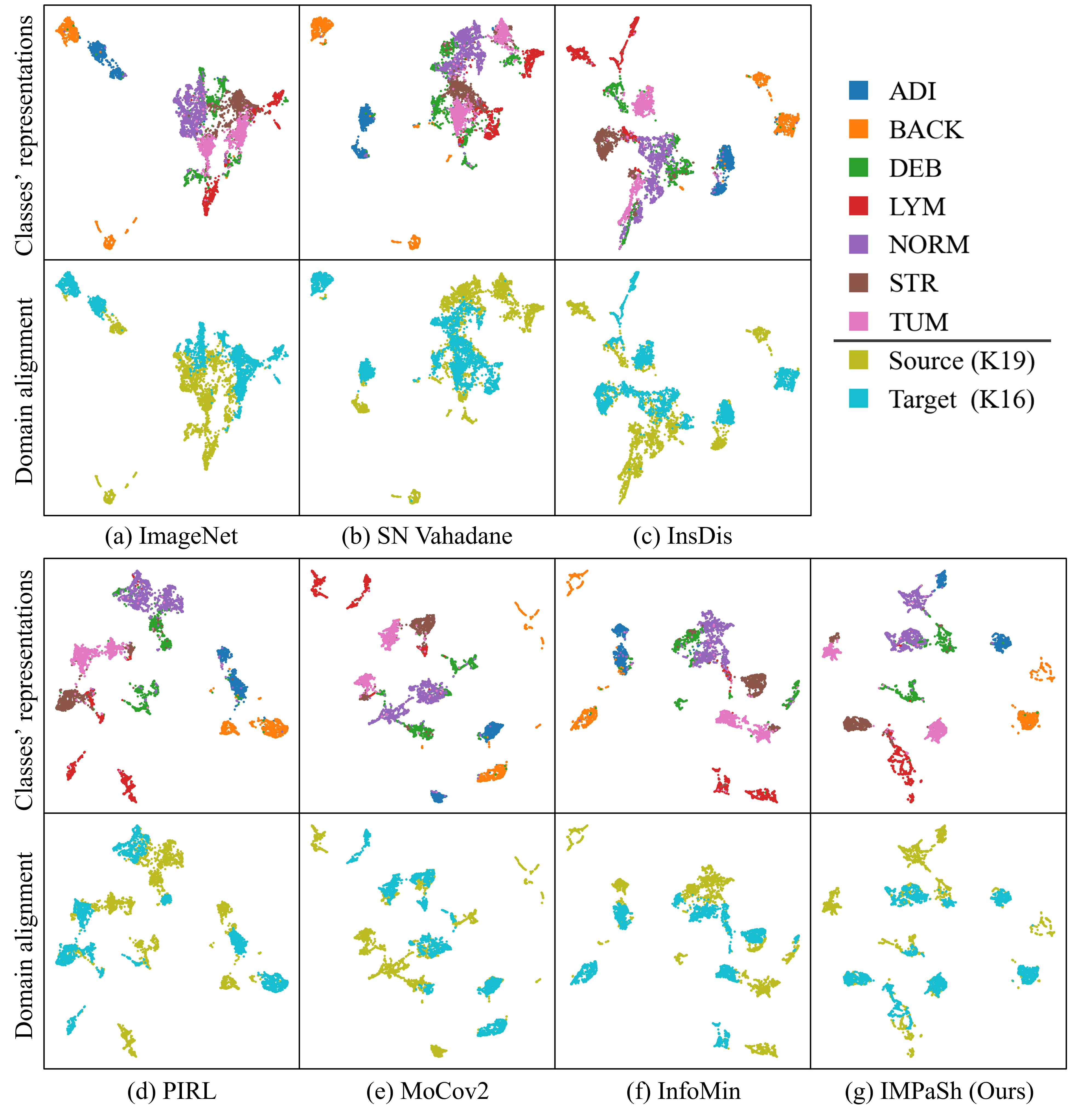}
\caption{The UMAP \cite{mcinnes2018umap} visualization of the source (K19) and target (K16) domain feature representation vectors for domain alignment (top row) and the different classes’ representations (bottom row). We compare our (g) IMPaSh methods and the a) baseline ImageNet encoder, b) ImageNet encoder+SN and 2 other self-supervised methods e) MoCoV2 and f) InfoMin.}
\label{umap}
\end{figure*}

\begin{table*}[!t]
\caption{Quantitative measurements on the degree of separations across clusters in term of class label and domain alignment for each method}\label{t:silhouette}

\begin{subtable}[t]{\textwidth}
\centering
\caption{Silhouette scores when measuring clusters of each class (higher is better)}
\begin{tabular}{c|c|c|c|c|c|c|c}
\toprule
\textbf{Domains} & ImageNet & \begin{tabular}{@{}c@{}}SN \\ Vahadane\end{tabular} & InsDis & PIRL & MoCov2 & InfoMin & IMPaSh \\
\midrule
Target & 0.402 & 0.461 & 0.465 & 0.485 & 0.410 & 0.471 & \textbf{0.553} \\
All & 0.144 & 0.199 & 0.246 & \textbf{0.360} & 0.171 & 0.264 & 0.311 \\
\bottomrule
\end{tabular}
\end{subtable}

\vspace*{2.5mm}
\centering

\begin{subtable}[!t]{\textwidth}
\centering
\caption{Silhouette scores when measuring clusters of each domain \textit{\textbf{within}} each class cluster (lower is better)}
\setlength\tabcolsep{6pt} 
\begin{tabular}{c|c|c|c|c|c|c|c}
\toprule
\textbf{Class} & ImageNet & \begin{tabular}{@{}c@{}}SN \\ Vahadane\end{tabular} & InsDis & PIRL & MoCov2 & InfoMin & IMPaSh \\
\midrule
ADI & 0.557 & 0.559 & 0.497 & 0.580 & 0.689 & 0.544 & 0.667 \\
BACK & 0.808 & 0.671 & 0.617 & 0.614 & 0.687 & 0.707 & 0.580 \\
DEB & 0.214 & 0.139 & 0.137 & 0.132 & 0.067 & 0.056 & 0.057 \\
LYM & 0.815 & 0.719 & 0.652 & 0.517 & 0.500 & 0.436 & 0.332 \\
NORM & 0.304 & 0.181 & 0.282 & 0.291 & 0.335 & 0.384 & 0.320 \\
STR & 0.448 & 0.499 & 0.405 & 0.418 & 0.413 & 0.442 & 0.418 \\
TUM & 0.647 & 0.512 & 0.448 & 0.426 & 0.582 & 0.580 & 0.545 \\
\midrule
All & 0.542 & 0.469 & 0.434 & 0.425 & 0.468 & 0.450 & \textbf{0.417} \\
\bottomrule
\end{tabular}
\end{subtable}

\end{table*}

We conducted an ablation study and reported the results in Table \ref{tab2}. The results suggest that PatchShuffling provided better performance compared to PIRL when used in conjunction with additional projection heads and momentum branches. On the other hand, using momentum encoders can be detrimental to model overall performance when jointly utilizing InfoMin and PIRL.

To qualitatively assess the impact of each technique on the actual image representation, we utilized UMAP (Uniform Manifold Approximation and Projection) \cite{mcinnes2018umap} for visualizing the distribution of samples (i.e., their ResNet50 features) from both source (K19) and target domain (K16) in  Fig. \ref{umap}. For NORM (purple) and STR (brown) classes, we observed that the top three methods (PIRL, InfoMin and our proposed IMPaSh) were able to noticeably increase the distance between these two clusters while keeping the samples from the source (lime green) and target domains (cyan) that belong to each of these clusters close to each other. On the other hand, image features obtained using methods with less competitive results like the baseline or stain normalization are highly clumped together or visibly closer. These observations suggest that the features from IMPaSh are highly resistant when switching the domain from K19 to K16 for NORM and STR. In the same vein, IMPaSh seems to provide better representation compared to PIRL and InfoMin on LYM (red), given the small area occupied by their samples. Nonetheless, all methods have little success in making NORM (purple) and DEB (green) categories more distinguishable. In our case, while IMPaSh features for NORM and DEB samples are nicely separated into clusters that are far from each other, the samples from the source (lime green) and target (cyan) domains for the same label are not close, thus indicating a strong shift in distribution.

In addition to the qualitative results, we further utilized silhouette scores to measure the degree of separation of the clusters obtained from each method in terms of the classes and the domain alignment. Intuitively, from a better domain generalization technique, we would be able to obtain clusters that satisfy: a) the clusters of each class across all domains will get further apart from each other while b) the clusters of each domain \textit{within} each class cluster get pulled closer. In other words, a better method has higher silhouette scores when measuring the clusters of class labels across domains (class-level scores). At the same time, for methods that have close class-level scores, one that has a smaller silhouette score when measuring for domain alignment within each class label (domain-level score) is the better. We present the class-level scores in Table \ref{t:silhouette}a and domain-level scores in Table \ref{t:silhouette}b. Consistent with our intuition, we observe that methods that have better performance in Table \ref{tab1} achieved higher scores than others in Table \ref{t:silhouette}a. Interestingly, PIRL has a higher score than IMPaSh while having noticeably lower classification performance. When measuring domain-level separability in \ref{t:silhouette}b, IMPaSh achieved the lowest score when averaging across all classes, but PIRL is the closest in terms of overall performance. All in all, results in \ref{t:silhouette}a and \ref{t:silhouette}b further demonstrate the benefits of IMPaSh framework. However, contradictory observations for PIRL and IMPaSh suggest further investigations in the future are required.

\section{Conclusion}

We proposed a new augmentation named PatchShuffling and a new SSCL framework called IMPaSh that utilize momentum, PatchShuffling and InfoMin to pre-train a neural network encoder. We demonstrated that the resulting encoder was able to extract discriminative image representation while being highly robust against distribution shift. However, our research stops short of evaluating only colorectal tissue classification. Further investigations are required to identify if our IMPaSh can scale for other tasks and/or with more data. In addition, as PatchShuffling is quite modular, investigations on using it in combination with other domain-specific augmentations may also be beneficial for improving IMPaSh.

\section*{Acknowledgements} This work was funded by the Medical Research Council (MRC) UK and South Korea biomedical and health researcher exchange scheme grant No. MC/PC/210-14, and the National Research Foundation of Korea (NRF) grant funded by the Korean government (MSIP) (No. 2021K1A3A1A88100920 and No. 2021R1A2C2-014557).

\bibliographystyle{IEEEtran}
\bibliography{egbib.bib}

\end{document}